\pdfoutput=1
\documentclass[10pt,letterpaper]{article}

\usepackage{surveillancebench-preprint}
\usepackage{colortbl}

\preprintTitle{BioSecBench-Surveillance}
\preprintSubtitle{A Verifiable Benchmark for AI Agents in Pathogen Genomic Surveillance}

\preprintAuthors{Harmon Bhasin\preprintSup{1,*}, Kevin Flyangolts\preprintSup{2,*}, Dianzhuo Wang\preprintSup{1}, Evan Seeyave\preprintSup{1}, Arjun Banerjee\preprintSup{1}, Amanda Darling, Joshua Stallings, David Stern, Shawn Higdon, Claire Duvallet, Bryan Tegomoh, Kenny Workman\preprintSup{1}}

\preprintAffiliations{\preprintSup{1}LatchBio, San Francisco, CA \quad \preprintSup{2}Aclid, New York, NY \\[2pt] \\[2pt] \preprintSup{*}Harmon Bhasin and Kevin Flyangolts contributed equally.}

\preprintCorrespondence{kenny@latch.bio}

\newcommand{\nEvals}{100}
\newcommand{\nConfigs}{sixteen}
\newcommand{\nSampleTypes}{six}
\newcommand{\nTaskCats}{seven}

\begin{document}

\PreprintHeader

\begin{PreprintAbstract}
As pathogen genomic surveillance scales, the bottleneck is shifting from data generation to analysis. We present BioSecBench-Surveillance, a verifiable benchmark of \nEvals{} evaluations testing whether AI agents can infer the right analysis pipeline from raw sequencing data and surveillance context. Each evaluation gives an agent only the data and context a human analyst would have, then grades its structured answer deterministically. The tasks span \nTaskCats{} categories, from taxonomic classification to genetic-engineering detection, across diverse sample types and sequencing technologies. Across 3,962 gradable attempts from \nConfigs{} model–harness pairs, the strongest configuration cleared only about half. Opus 4.8 / PI led at 50.2\% (95\% confidence interval (CI), 40.1–60.3; 83 evaluations), tied with GPT-5.5 / Codex at 50.2\% (95\% CI, 40.8–59.6), followed by Opus 4.7 / PI at 49.6\% (95\% CI, 40.0–59.2) and Sonnet 4.6 / PI at 48.6\% (95\% CI, 38.9–58.3). Even when agents invoked the correct workflows, their mistakes came from the choices around them, such as which references, thresholds, filters, and normalization to apply. BioSecBench-Surveillance provides a standard for measuring whether agents can be trusted to perform genomic surveillance when the next outbreak arrives.

\end{PreprintAbstract}

\StartBody

\section{Introduction}

Genomic surveillance is a central instrument in pandemic preparedness and public health \cite{ref_gardy_loman}. Sequencing clinical, agricultural, environmental, and wastewater samples can identify pathogens, resolve transmission, and detect emerging threats at a resolution that culture- and antigen-based methods often lack \cite{ref_gardy_loman,ref_nejm_pathogen_genomics}. Falling sequencing costs and expanding programs keep increasing the volume and breadth of these data \cite{ref_who_genomic_strategy}. However, these data inform preparedness only once analyzed, and that analysis remains expert-driven and difficult to scale. The bottleneck in preparedness is shifting from generating data to interpreting it \cite{afolayan2021overcoming}.

AI agents are a promising solution, as they can analyze complex data efficiently; however, it is unclear whether they can perform the right analysis for genomic surveillance, where that choice depends on many interacting variables. An analyst must select references, databases, filters, assembly or alignment strategies, lineage and typing systems, abundance methods, and thresholds, each depending on the organism, assay, sampling context, and access constraints. These choices are consequential: wastewater sequencing has revealed cryptic SARS-CoV-2 transmission before or outside clinical sequencing \cite{ref_karthikeyan_wastewater} and triggered outbreak investigation and vaccination response after a vaccine-derived poliovirus was linked to a paralytic case \cite{ref_cdc_polio_wastewater}. In both cases, the result depended on separating a real viral signal from a mixed environmental background and interpreting it against the appropriate references and controls.

These analytical choices make pathogen genomic surveillance a natural test case for execution-based benchmarking, which grades not whether an agent runs an analysis but whether it runs the right one. This kind of evaluation was popularized by SWE-bench \cite{ref_swebench}, which scores agents on whether their code changes resolve real GitHub issues under executable tests, rather than static question answering. Recent biology benchmarks extend them to scientific workflows in domains such as epigenetics and therapeutics, handing agents realistic workflow snapshots and grading structured responses deterministically \cite{ref_epibench,ref_txbench}. However, none covers the core tasks of pathogen surveillance; the closest one tests only whether an agent can retrieve relevant records from public sequence databases \cite{ref_gget_virus}.

We present BioSecBench-Surveillance, a verifiable benchmark that measures how well AI agents perform on pathogen genomic surveillance tasks. Each of its \nEvals{} evaluations snapshots a workflow at the moment before a surveillance decision, giving the agent only the context and files that a human analyst would have. The agent must decide which analysis to run, with which tools, databases, references, and thresholds, then return a structured answer graded deterministically. The evaluations cover \nTaskCats{} categories of surveillance analysis, from taxonomic classification to genetic-engineering detection, across \nSampleTypes{} sample types and both short- and long-read sequencing.

Current agents are not yet reliable for these tasks. In the \nConfigs{} model-harness configurations we tested, agents answered from $\sim$14\% to $\sim$50\% of gradable tasks correctly. When they answered incorrectly, they usually ran the right workflows but made the wrong choices around them: the wrong references, thresholds, or normalization. We release BioSecBench-Surveillance as a diagnostic for building agents that can be trusted to turn surveillance data into decisions for future outbreaks.

\EndBody


\StartBody

\section{Benchmark construction}

Each evaluation is built around a single empirical decision drawn from a real biosurveillance workflow. The task specifies the target decision and the required response; however, it does not mention the correct workflow. The input is raw or near-raw sequencing data along with any useful reference files. The ground truth of each answer is derived with validated gold-standard workflows, following published methods and literature. Each agent response is scored by a typed deterministic grader, and every evaluation is internally peer-reviewed by domain experts before inclusion. See \hyperref[sec:methods]{Methods} for more details.



\subsection{Evaluation inventory}

We break down BioSecBench-Surveillance into seven categories of tasks found in pathogen genomic surveillance (Figure~\ref{fig:composition}):

\begin{itemize}[leftmargin=1.2em,itemsep=1pt,topsep=2pt,parsep=0pt]
  \item \textbf{Variant detection}: assigning lineages, clades, or within-species sequence changes.
  \item \textbf{Taxonomic classification}: determining composition of a sample and quantifying abundance.
  \item \textbf{Antimicrobial resistance (AMR) characterization}: detecting resistance genes or mutations.
  \item \textbf{Source tracking}: attributing a sample to a likely host, source, or origin.
  \item \textbf{Toxin and virulence characterization}: identifying markers of toxin production or virulence.
  \item \textbf{Genetic-engineering characterization}: assessing whether sequence evidence is consistent with engineering or sample construction.
  \item \textbf{Anomaly detection}: flagging unexpected taxa, abundance profiles, or out-of-background signals with no prescribed target.
\end{itemize}

We grade only the final answer of each task, but reaching it requires many intermediate, ungraded choices about references, thresholds, filters, and controls. Given the uneven distribution of tasks across labels, we report per-label counts with every breakdown.

\EndBody

\begin{center}
\vspace{0.15in}
\includegraphics[width=0.95\textwidth]{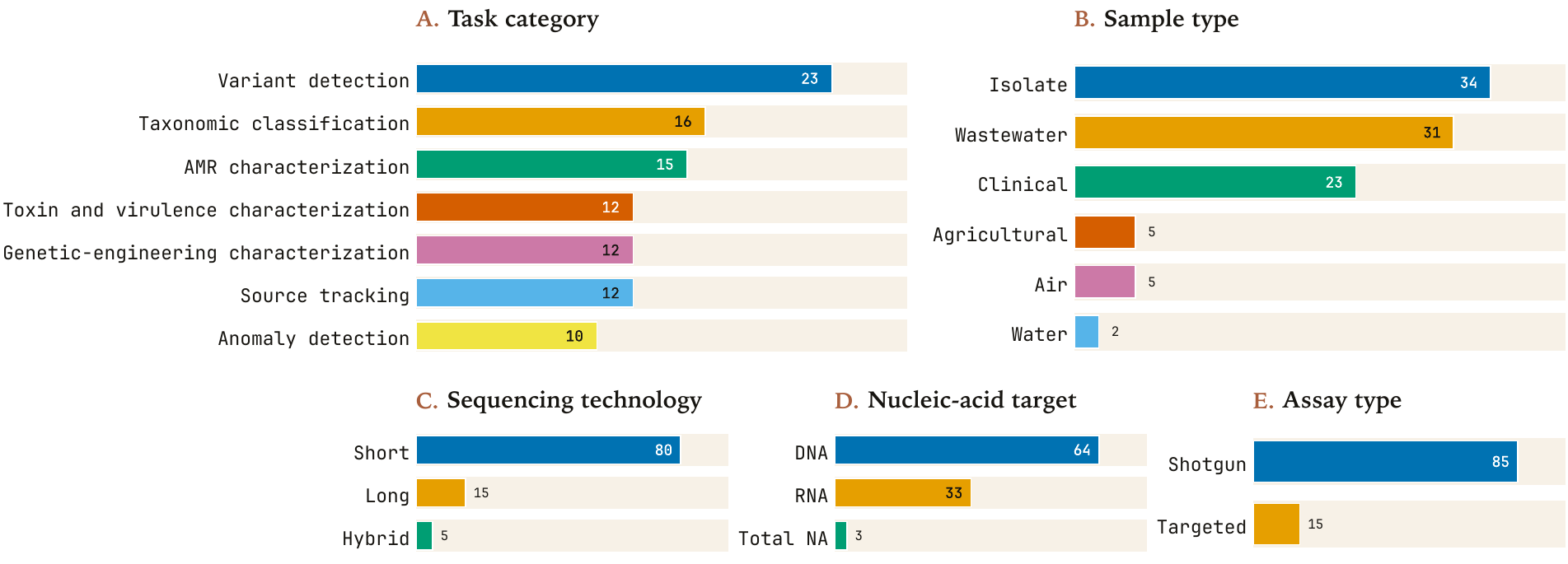}
\captionof{figure}{\textbf{Composition of the \nEvals{} BioSecBench-Surveillance evaluations.} Ranked counts of evaluations by (A)~task category, the biosurveillance decision under test; (B)~sample type, the specimen the sequencing data derive from; (C)~sequencing technology (short-read, long-read, or hybrid); (D)~nucleic-acid extraction target (DNA, RNA, or total nucleic acid, ``Total NA''); and (E)~assay type (shotgun versus targeted/amplicon/capture). Bar labels give evaluation counts.}
\label{fig:composition}

\end{center}

\StartBody

\section{Results}

\subsection{No model–harness pairing exceeds $\sim$50\% mean pass rate}

The endpoint pass rates (correct answers as a share of gradable attempts) ranged from approximately 14\% to 50\% in the sixteen configurations and averaged 41\% (Figure~\ref{fig:topline}B). Each configuration pairs one model with one inference harness, the agent scaffold (Claude Code, PI, or Codex) that gives the model its tools and prompt structure. Only the two xAI configurations clearly separated, trailing 14 to 16\%; the Anthropic, OpenAI, and Google configurations overlapped between 38\% and 50\%. Even the two best configurations, Opus 4.8 with PI and GPT-5.5 with Codex, passed only about half of the gradable attempts.

Refusals, a third outcome in which the request is declined rather than analyzed, varied enormously between configurations, from none at all to nearly a third of tasks (Figure~\ref{fig:topline}A). OpenAI configurations refused 27 to 29\% of tasks under PI but only 8 to 9\% of the same tasks under Codex. Anthropic configurations refused 18 to 31\% and Google 7 to 15\%, while xAI configurations never refused.


\EndBody

\begin{center}
\includegraphics[width=0.95\textwidth]{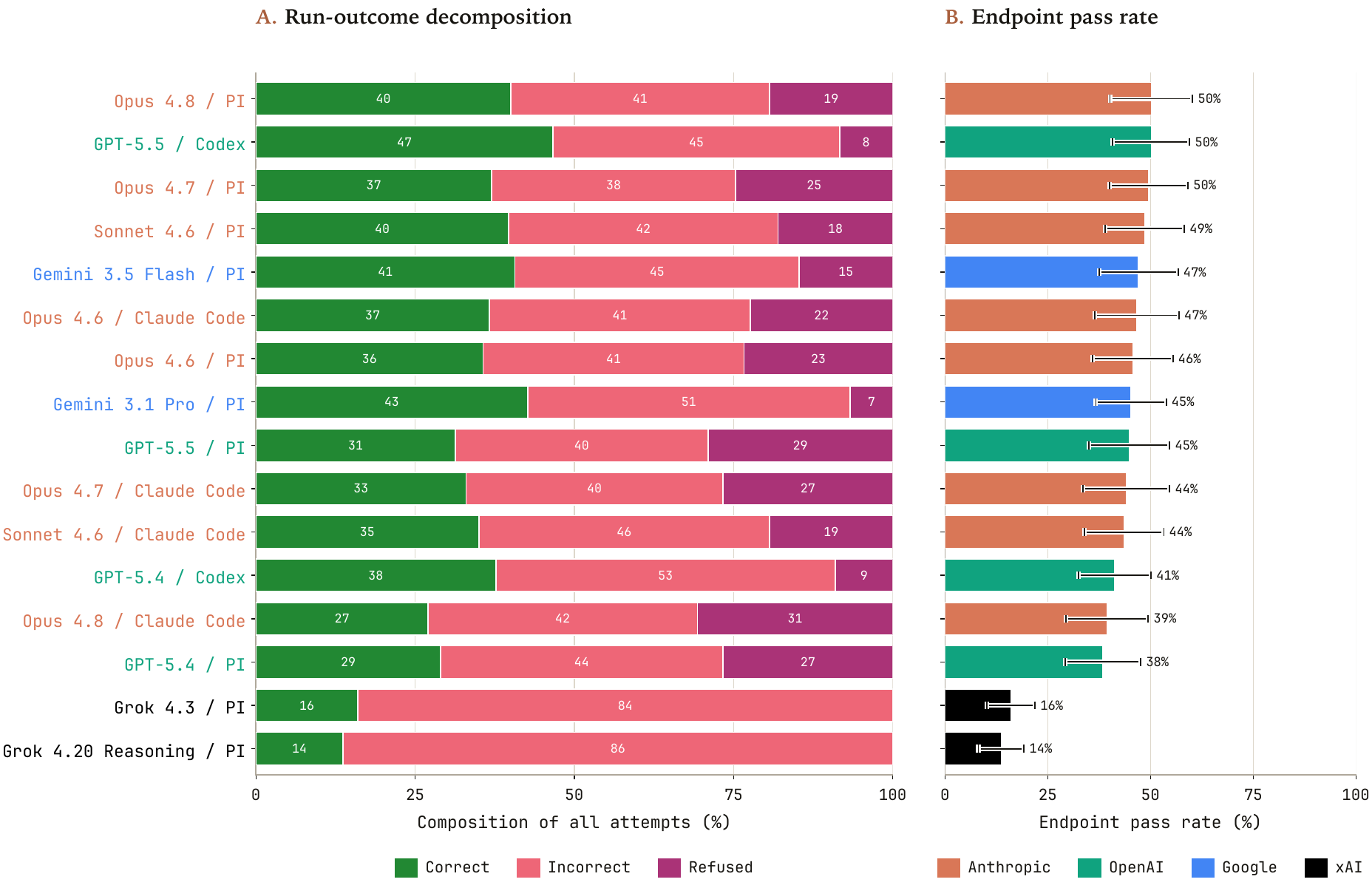}
\captionof{figure}{\textbf{Topline BioSecBench-Surveillance performance across \nConfigs{} model~$\times$~harness configurations.} Configurations are ordered by endpoint pass rate. (A)~Run-outcome decomposition: every attempt is correct, incorrect (a wrong answer, a timeout, or an ungradable answer), or refused, shown as a share of all attempts. (B)~Endpoint pass rate per configuration, the mean of per-evaluation pass rates (correct answers as a share of gradable attempts, i.e., those not refused), colored by model provider; error bars are 95\% $t$ intervals over evaluations.}
\label{fig:topline}
\end{center}

\StartBody

\subsection{Difficulty varies most with task and read technology, and least with sample type and target}

Most task categories and sample types scored between 35\% and 50\%, with anomaly detection being the lowest by a wide margin (Figure~\ref{fig:breakdown}). Among the task categories (Figure~\ref{fig:breakdown}A), six of the seven clustered in this band, led by source tracking and taxonomic classification (50\% and 46\%), while anomaly detection fell to 20\%, well below genetic-engineering characterization, the next hardest at 35\%. Sample types varied less (Figure~\ref{fig:breakdown}B): clinical (45\%) and isolate (44\%) specimens were handled best, and wastewater was somewhat worse (37\%).

Among sequencing attributes, read length mattered most: long-read datasets were the hardest at 26\%, below short-read (41\%) (Figure~\ref{fig:breakdown}C). The nucleic-acid target and the assay type moved far less: DNA and RNA differed only modestly (39\% and 42\%), and shotgun and targeted assays were indistinguishable (about 40\% each) (Figure~\ref{fig:breakdown}D–E).



\EndBody

\begin{center}
\includegraphics[width=0.95\textwidth]{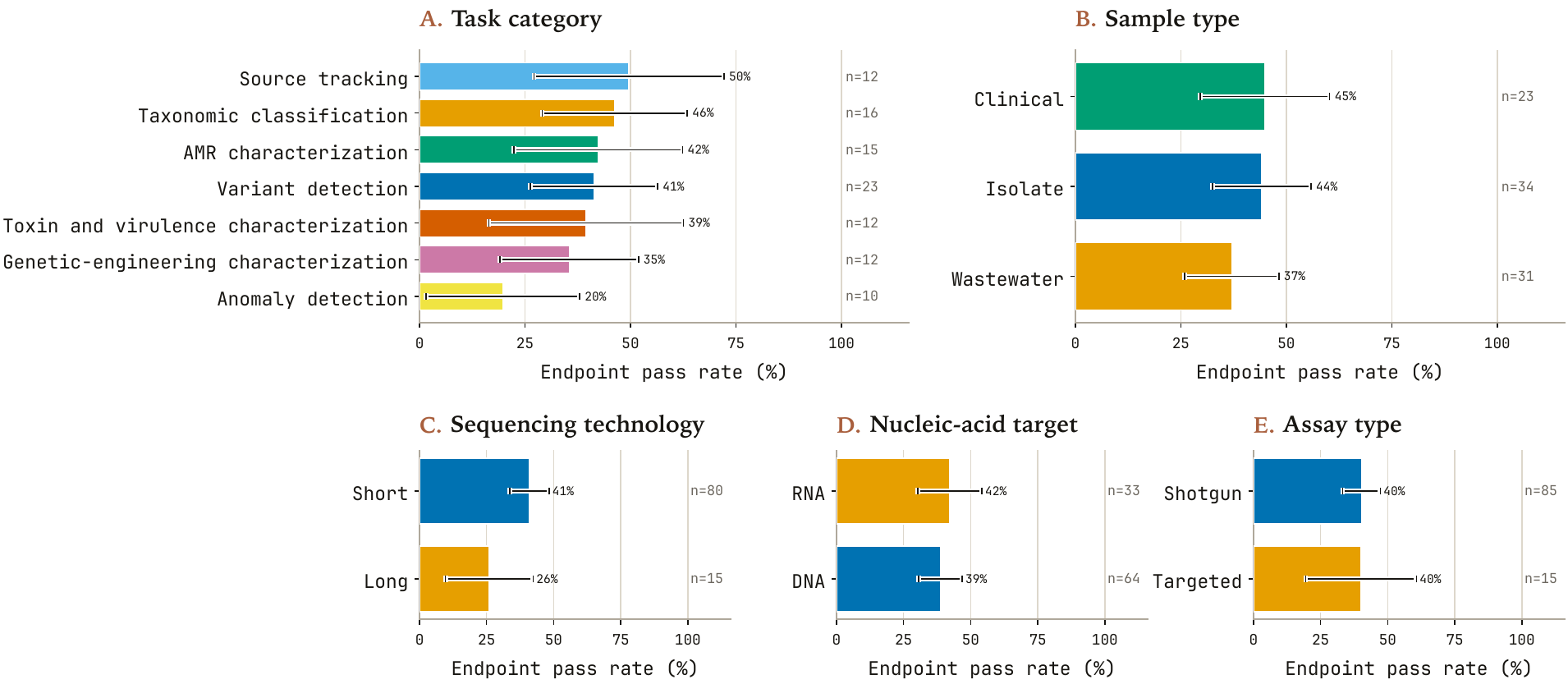}
\captionof{figure}{\textbf{BioSecBench-Surveillance performance broken down by evaluation attributes.} Endpoint pass rate by (A)~task category, (B)~sample type, (C)~sequencing technology, (D)~nucleic-acid target, and (E)~assay type. Each evaluation was reduced to a one pass-rate estimate averaged across the \nConfigs{} configurations. Bar labels give the pass rate; \texttt{n} is the number of evaluations, and cells with fewer than 10 evaluations are omitted. Error bars are 95\% $t$ intervals over evaluations.}
\label{fig:breakdown}
\end{center}

\StartBody

\subsection{Failure patterns are consistent across configurations}

Configurations differ in overall pass rate but not in where or how they fail. The category difficulty ranking is largely conserved (Figure~\ref{fig:models}): anomaly detection and genetic-engineering characterization are the hardest for nearly every configuration, while source tracking, taxonomic classification, and variant detection are easier across the board. Trajectory review finds the same failure signature across providers: agents almost always call the validated tools rather than re-implementing them, and their errors come instead from the choices around those tools (reference or database, thresholds and filters, normalization), which can leave a plausible answer outside the grading tolerance.


\EndBody

\begin{center}
\includegraphics[width=0.95\textwidth]{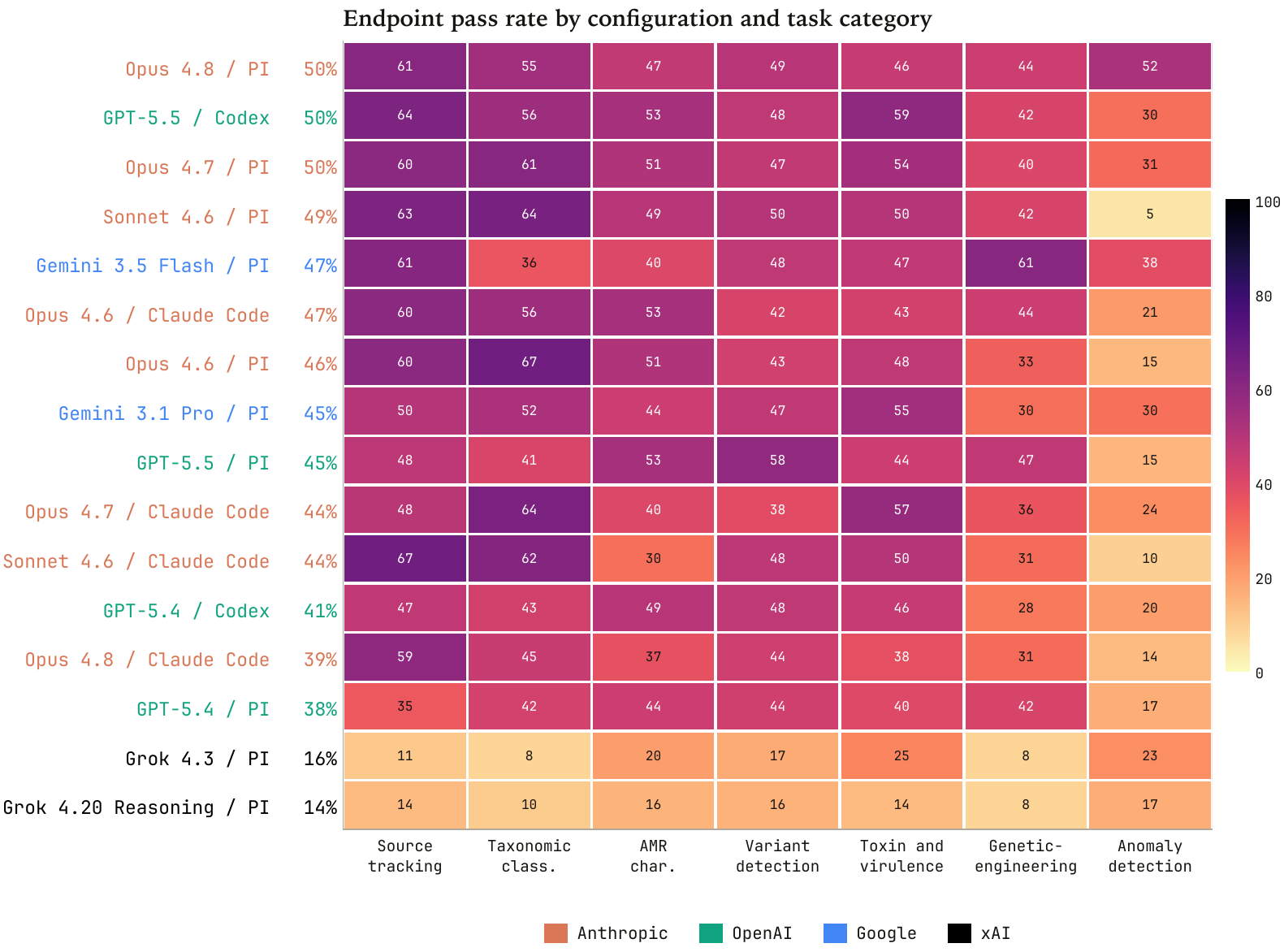}
\captionof{figure}{\textbf{Endpoint pass rate by configuration and task category.} Each cell is the endpoint mean pass rate for one model-harness configuration on one task category; darker is a higher pass rate. Rows are configurations ordered by overall pass rate (shown beside each label; strongest at top) and colored by provider. Columns are task categories ordered by overall pass rate (easiest at left).}
\label{fig:models}
\end{center}

\StartBody

\section{Discussion}

BioSecBench-Surveillance measures whether AI agents can make the analytical decisions required in pathogen genomic surveillance. Across every configuration we tested, they cannot do so reliably: none answered more than about half of the tasks correctly. In trajectory review, agents almost always reached for the right tools, then undercut them with the wrong references, thresholds, or normalization.

Beyond misconfigured tools, a harder failure appeared wherever a task turned on interpretation: the agents ran the correct analysis but drew the wrong conclusion from its output. This is what made anomaly detection (20\%) and genetic-engineering characterization (35\%) the two hardest categories, each an open-world call with no prescribed target. On anomaly detection they leaned too conservative: shown a low-abundance but high-consequence organism, they often dismissed it as environmental background and only sometimes flagged the wrong taxon. On engineered-sequence detection they could not reliably tell a deliberate construct from native or homologous sequence: most often they detected the engineering but misidentified its type or location, and less often either missed it entirely or, more rarely, called a native plasmid engineered.

The failures we describe are only meaningful because the tasks force agents to actually run the analysis, which require stripping every identifying clue from the inputs. Left in, a filename, a sequence header, or a stray accession was often enough for an agent to name the organism and report the expected answer without running the analysis at all. We flag this as a design requirement for anyone building benchmarks of this kind.

Several limitations of the benchmark suggest directions for future work. Tasks are unevenly distributed across categories, sample types, and sequencing technologies, so aggregate scores weight the better-populated labels more heavily; we report per-label counts throughout, but a larger and more balanced task set would allow the scores to be read directly. Refusals present a subtler problem, since a refused task has no gradable attempt and drops out of a configuration's denominator, leaving the configurations that refuse most scored on the fewest and least comparable tasks. This also opens a way to game the score: a model could raise its pass rate by declining the tasks it is most likely to fail. None of the frontier models we tested show this pattern, but pairing the benchmark with a dedicated refusal evaluation \cite{wintermute2026evaluatingcalibratedrefusalsafe} would protect against a model that does. Most fundamentally, we limited the benchmark to tasks with a single objective answer so they could be graded deterministically; yet much of real biosurveillance turns on judgment calls with no one agreed-upon method or answer. Those analyses are beyond what we can grade today, and measuring that open-ended judgment, as agents improve and outgrow benchmarks like this one, is the central problem the next generation of benchmarks will have to solve.

The gap between today's agents and a trusted analyst is judgment: which reference to trust, which threshold to set, which call to make when the signal is ambiguous. BioSecBench-Surveillance isolates that gap and grades it deterministically, the first step toward agents we can rely on when the next outbreak arrives.

\section{Methods}\label{sec:methods}

\subsection{Benchmark composition and data}

Each evaluation is a single definition file (a task prompt, a grader, metadata tags, structured notes, and pointers to the input data), and these files are the benchmark's source of truth. The \nEvals{} evaluations span \nTaskCats{} task categories and \nSampleTypes{} sample types. The input data are raw or near-raw sequencing artifacts (FASTQ reads, assembled contigs, or aligned reads) drawn either from real National Center for Biotechnology Information (NCBI) Sequence Read Archive (SRA) datasets or from simulated reads constructed to encode a known ground truth. Any identifying metadata is removed from the files. The ground truth is established with gold-standard workflows or literature and is kept out of the prompt. Each ground truth is designed to be robust to tooling changes as long as the general correct workflow is attempted. For each evaluation, the notes section documents why the question was chosen, what the optimal workflow to get the answer might look like, along with other information, and is internally peer-reviewed for technical accuracy before inclusion.

\subsection{Task format and deterministic grading}

The primary benchmark score is a deterministic endpoint grading of the final structured answer. Each task specifies the target decision and the exact answer schema. The agent then returns its answer as JSON which is written as a file in the workspace and analyzed by a grader. Each grader is built from typed field checks: numeric fields against a relative, absolute, minimum, or maximum tolerance; set-valued fields by label overlap above a Jaccard threshold; and categorical fields by exact or normalized match, multiple choice, or per-key dictionary comparison. When an answer spans several fields of different kinds, these checks are combined by an explicit rule, for example an all-of node that passes only if every field passes. Tolerances are set per evaluation from the underlying biology and how the ground truth was derived, and are recorded in the notes. A run passes only when all required checks pass; missing fields, invalid JSON, off-schema answers, and answers the grader cannot parse count as failures.


\subsection{Agent runs and execution}

Each model-harness pair was run three times per evaluation on the LatchBio data
infrastructure. A configuration is a model paired with an inference harness,
the scaffold (Claude Code, PI, or OpenAI Codex) that gives the model its tools,
file access, and prompt and turn structure; we treat model and harness as
independent axes because the same model behaves very differently across them,
in both capability and refusal. We evaluated \nConfigs{} deployed pairs: Opus
4.8, 4.7, and 4.6 and Sonnet 4.6 under Claude Code (CC) and under PI, GPT-5.5
and 5.4 under OpenAI Codex and under PI, Grok 4.3 and 4.20 under PI, and Gemini
3.5 Flash and 3.1 Pro under PI. Each run executed in an identical containerized
sandbox preloaded with a standard suite of open-source bioinformatics tools,
including assemblers, aligners, taxonomic classifiers, variant callers, and
antimicrobial-resistance and virulence databases, so each agent chooses among
the same tools a human analyst would use. Each sandbox provides six CPU cores,
34~GiB of memory, 512~GiB of disk, and a six-hour wall-clock limit. The sandbox
permits internet access so an agent can fetch additional references or
databases at run time. The evaluation's data files are staged into the
workspace so the agent can read them, and the harness records a complete raw
trajectory (the conversation, tool calls, and execution outputs) for every run.
The \nConfigs{} pairs were run over 3 trials of the \nEvals{} evaluations, for
roughly 4{,}800 runs in total.

\subsection{Outcome classification and aggregation}

Each run is assigned one of three mutually exclusive outcomes that sum to the run count for every evaluation and configuration: correct, incorrect, or refused. The endpoint pass rate is correct runs over the sum of correct and incorrect runs (excluding refusals). Incorrect runs also include the rare run that timed out without producing an answer; because experts using the correct methods finish well within the time limit, a timeout indicates the model took a wrong or inefficient path rather than one that ran out of time. All three outcomes are reported in the decomposition (Figure~\ref{fig:topline}A). Uncertainty intervals are 95\% Student-$t$ intervals computed across per-evaluation pass rates, with the evaluation as the sampling unit.

\section{Data availability}

A public subset of BioSecBench-Surveillance, comprising example evaluations with task prompts and descriptive metadata, is available at \href{https://github.com/latchbio/biosecbench-surveillance}{github.com/latchbio/biosecbench-surveillance}. In accordance with biosecurity research practice, the entire evaluation set is held under restricted access.

\EndBody
\clearpage
\bibliographystyle{unsrt}
\bibliography{references}

\end{document}